\pdfoutput=1

\documentclass[11pt]{article}

\usepackage{acl}

\usepackage{times}
\usepackage{latexsym}
\usepackage{booktabs}
\usepackage{multirow}
\usepackage{array}

\usepackage[T1]{fontenc}

\usepackage[utf8]{inputenc}
\usepackage{hyperref}

\usepackage{microtype}

\usepackage{inconsolata}

\usepackage{graphicx}
\usepackage{amsmath}
\usepackage{float}

\title{\texttt{FineZip} : Pushing the Limits of Large Language Models for \\ Practical Lossless Text Compression}

\author{Fazal Mittu$^1$, Yihuan Bu$^1$, Akshat Gupta$^1$, Ashok Devireddy$^1$, Alp Eren Ozdarendeli$^1$, \\
\textbf{Anant Singh$^2$, Gopala Anumanchipalli$^1$}\\
$^1$UC Berkeley, $^2$NYU\\
  \small{
   \href{akshat.gupta@berkeley.edu}{akshat.gupta@berkeley.edu}
  }
}

\begin{document}
\maketitle
\begin{abstract}
While the language modeling objective has been shown to be deeply connected with compression, it is surprising that modern LLMs are not employed in practical text compression systems. In this paper, we provide an in-depth analysis of neural network and transformer-based compression techniques to answer this question. We compare traditional text compression systems with neural network and LLM-based text compression methods. Although LLM-based systems significantly outperform conventional compression methods, they are highly impractical. Specifically, LLMZip, a recent text compression system using Llama3-8B requires 9.5 days to compress just 10 MB of text, although with huge improvements in compression ratios. To overcome this, we present \texttt{FineZip} - a novel LLM-based text compression system that combines ideas of online memorization and dynamic context to reduce the compression time immensely. \texttt{FineZip} can compress the above corpus in approximately 4 hours compared to 9.5 days, a 54 times improvement over LLMZip and comparable performance. \texttt{FineZip} outperforms traditional algorithmic compression methods with a large margin, improving compression ratios by approximately 50\%. With this work, we take the first step towards making lossless text compression with LLMs a reality. While  \texttt{FineZip} presents a significant step in that direction, LLMs are still not a viable solution for large-scale text compression. We hope our work paves the way for future research and innovation to solve this problem. 


\end{abstract}


\begin{figure*}[ht]
    \centering
    \includegraphics[width=0.9\textwidth]{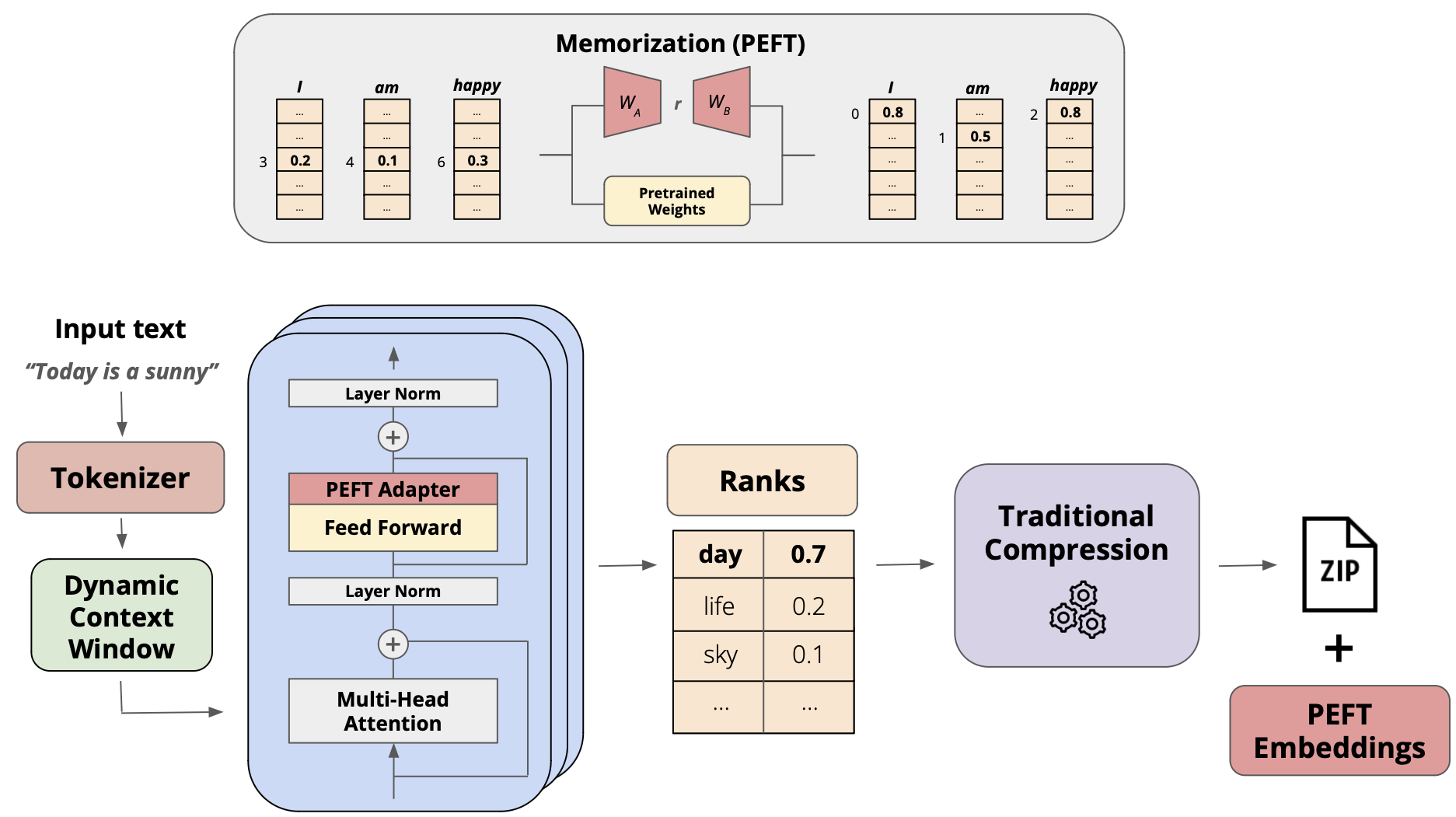} 
    \centering
    \caption{System diagram for \texttt{FineZip}.}
    \centering
    \label{fig:system}
    \centering
\end{figure*}

\section{Introduction}
While the relationship between language modeling and compression has long been known \citep{sequential, mahoney2000fast, goyal2018deepzip, Bellard2019LosslessDC}, recent works \citep{delétang2024language, huang2024compression} have reinforced this connection. \citet{delétang2024language} recently showed large language models (LLMs) can be used to compress data from various modalities. \citet{huang2024compression} followed up this work by showing that increasing compression abilities of LLMs is linearly correlated to downstream task performance. 


Previous works have exploited this connection for lossless text compression. Neural network based models have been implemented for text compression \citep{sequential,mahoney2000fast, goyal2018deepzip} and have reached better compression performance than traditional algorithmic compressors such as gzip. More recent methods have explored using LSTM and transformer models \citep{Bellard2019LosslessDC,Bellard2021NNCPVL}. These methods fall under the "online" compressors category, where a randomly initialized model is directly trained on the data being compressed. In this case, the model parameters also become part of the compression. A recent effort, LLMZip \citep{valmeekam2023llmzip}, tested the use of LLMs for lossless compression. Given an LLM's ability to predict the next token provided a fixed-length context window, a tokenized text can be stored as probabilistic ranks produced by an LLM predicting the next token. This is a type of "offline" compression, with a fixed system used for both compression and decompression of all incoming text.

In this paper, we build on prior work and introduce \texttt{FineZip}, which uses LLMs for lossless text compression with both online and offline components. \texttt{FineZip} combines an "online" component, which memorizes the data being compressed, with an "offline" component in the form of pre-trained LLMs for compression. The "online" memorization is done by fine-tuning the model on the data being compressed in a parameter-efficient way \citep{hu2021lora, dettmers2023qlora} with an additional constant overhead of the learned embeddings during fine-tuning. The "offline" component of the system is the pre-trained LLM which remains fixed across different corpora. Figure \ref{fig:system} depicts the system diagram for \texttt{FineZip}. With this approach, we can leverage the benefits of online compression for improved performance without the drawback of requiring additional storage for model parameters. 

Additionally, with \texttt{FineZip} we allow for a dynamic context where each token being compressed has a context size of equal to its position in a sentence. This allows us to batch compression and decompression steps using LLMs, allowing for significant speed-up. "Online memorization" using PEFT methods also allows the model to compensate for loss of performance due to a dynamic context, while a dynamic context allows for batching which allows compression and decompression of many batches of text in parallel within a fixed compute budget. With \texttt{FineZip}, we can achieve 54 times faster compression times with minor loss of performance when compared to LLMZip, still outperforming traditional text compression methods by a huge margin. Our work also shows that compression rates of LLM-based methods are still not low enough for practical use cases, and although \texttt{FineZip} pushes the limits of using LLMs lossless text compression in practice, much work still needs to be done. The code for our work can be found here - \url{https://github.com/fazalmittu/FineZip}.



\section{Introducing \texttt{FineZip}}

The most basic form of compression using LLMs would be to tokenize the input text. Since each character in a word occupies 8 bits (1 byte in UTF-8 encoding), representing the word as a token, essentially converting it into a number, will almost always reduce the number of bytes needed to represent it. This connection was also observed in \citet{delétang2024language}. As a next step, we can use the predictive capabilities of LLMs for compression. This idea is used in LLMZip \cite{valmeekam2023llmzip} where they use a pre-trained LLM for text compression. The connection between language modeling and compression becomes intuitive when we take a deeper look at the language modeling objective, implemented using a cross-entropy loss. It aims to make each token in the training data the most probable token given the context preceding it, thus minimizing the number of bits required to represent the rank of the token in the vocabulary list, when ranked in descending order according to their probability.  Following this line of thought, we propose an intuitive yet effective way of enhancing this - fine-tuning the model on the data being compressed.




A challenge towards fine-tuning modern LLMs is that they are memory-intensive. Additionally, if we fine-tune the entire model on the text being compressed, then the entire LLM becomes part of the compression, requiring an additional space equal to the space required to store the model for decompression. Thus, we propose \texttt{FineZip}, a compression framework that involves parameter-efficient fine-tuning (PEFT) \citep{peft} on the input text as an "online" step prior to compression. We call this the "online memorization" step which makes the data being compressed more probable for the LLM. This fine-tuning is implemented using LoRA \cite{hu2021lora} and is much faster than full fine-tuning, requires much less GPU memory, and requires a very small amount of additional storage for the trained embeddings. The additional embedding storage does not scale with the dataset being compressed and becomes negligible at large sizes of corpora. 





Another key difference between LLMZip and \texttt{FineZip} is that \texttt{FineZip} adopts a dynamic context size approach rather than maintaining a fixed sliding window. LLMZip uses a permanent sliding window approach, where the rank of each token produced has a fixed context window of a preset context size (512 as chosen by original authors). This by design makes the compression process extremely autoregressive and non-parallelizable, as to produce the rank of a token, you need the previous 512 tokens. 

\vspace{0.2cm}

\noindent \texttt{FineZip} overcomes this limitation by employing a two-step dynamic context window technique:
\begin{enumerate}
    \item Divide the corpus into chunks of a pre-decided window length.
    \item Produce the ranks of each token within the window such that the rank for the $i^{th}$ token is produced based on the tokens preceding it
\end{enumerate}

The dynamic context window gives a variable context size to each token in a chunk. For a uniform comparison, we use a chunking size of 512 in \texttt{FineZip}, which is the same as the context window size chosen by LLMZip. In \texttt{FineZip}, the $i^{th}$ token in a chunk has a context size of $i-1$, thus only the final token in a chunk has access to full context length of 512. In contrast, every token in LLMZip has access to the full context length of 512. The dynamic context leads to some loss of performance, which is made up for by online memorization.

\section{Experiments}
We begin by comparing \texttt{FineZip} with (i) traditional text compression methods - bzip2 \cite{bzip2}, zlib \cite{zlib}, and gzip \cite{gzip}, (ii) neural network based text compression methods - NNCP \cite{Bellard2021NNCPVL}, and the (iii) recent LLM-based text compression method called LLMZip. For both  \texttt{FineZip} and LLMZip, we use Llama-3 8B \cite{llama3}.

\paragraph{Modifications to LLMZip:} LLMZip originally used Llama-1-7B \cite{llama1} while we leverage Llama-3-8B for both LLMZip and \texttt{FineZip} for uniform comparison. Additionally, LLMZip used two methods for compression - one using arithmetic coding (AC) and the other using a secondary compression methods on generated ranks. LLMZip uses zlib \cite{zlib} as a secondary compression method over ranks whereas our experiments show that bzip2 provides a much better compression ratio (Appendix: \ref{sec:traditional}). Thus, we use bzip2 as our secondary compression method for LLM ranks in both LLMZip and \texttt{FineZip}. We also refer to bzip2 as the baseline for text compression using traditional compression methods (Table \ref{tab:compression_methods}). To offer a better comparison, we also create a version of \texttt{FineZip} that incorporates arithmetic coding. The process uses the logits that the LLM outputs for each new token as the probability distribution update for the arithmetic coding scheme.

\begin{table}
\small
\centering
\begin{tabular}{|c|c|c|}
\hline
\textbf{Method} & \textbf{Compression Ratio} & \textbf{Time (min)} \\ \hline
zlib& 0.3251&0.0083\\\hline
gzip& 0.3238&0.0141\\\hline
bzip2& 0.2374&0.0437\\\hline
NNCP & 0.15021 & 251 \\ \hline
LLMZip (AC)& 0.0795& 13571\\ \hline  
LLMZip & 0.1163& 13651\\ \hline
\texttt{Finezip} (AC) & 0.0797&13118\\\hline
\textbf{\texttt{Finezip}} & \textbf{0.12799} & \textbf{250} \\\hline
\texttt{Finezip}-4bit & 0.1445 & 67 \\\hline

\end{tabular}
\caption{Comparison of Compression Methods on 10mb}
\label{tab:compression_methods}
\end{table}

We used the first 10mb of the enwik8 \cite{enwik8} dataset which is a standard benchmark for compression tasks. Though compression ratio (ratio of compressed file size and original file size) is the key metric, we are also interested in measuring time taken by these compression systems to evaluate practicality. The results are shown in Table \ref{tab:compression_methods}. The first key observation is that neural network and LLM based compression methods have significantly better compression ratios than traditional text compression methods (zlib, gzip, bzip2), thus highlighting the potential impact of these methods for text compression. The second key observation is that neural network and LLM based methods takes a long time to compress even small amounts of text, thus preventing their use in practice. This is especially true when using AC for compression in LLM-based methods, which produces exceptional compression ratios but also requires unprecedentedly large amounts of time. For LLMZip with AC, the time taken to compress 10MB of data is approximately 9.5 days. Thus, we do not explore AC-based LLM compression further and strictly compare only rank-based LLM baselines.

\setlength{\belowcaptionskip}{-10pt}
\begin{figure}
    \centering
    \includegraphics[width=\linewidth]{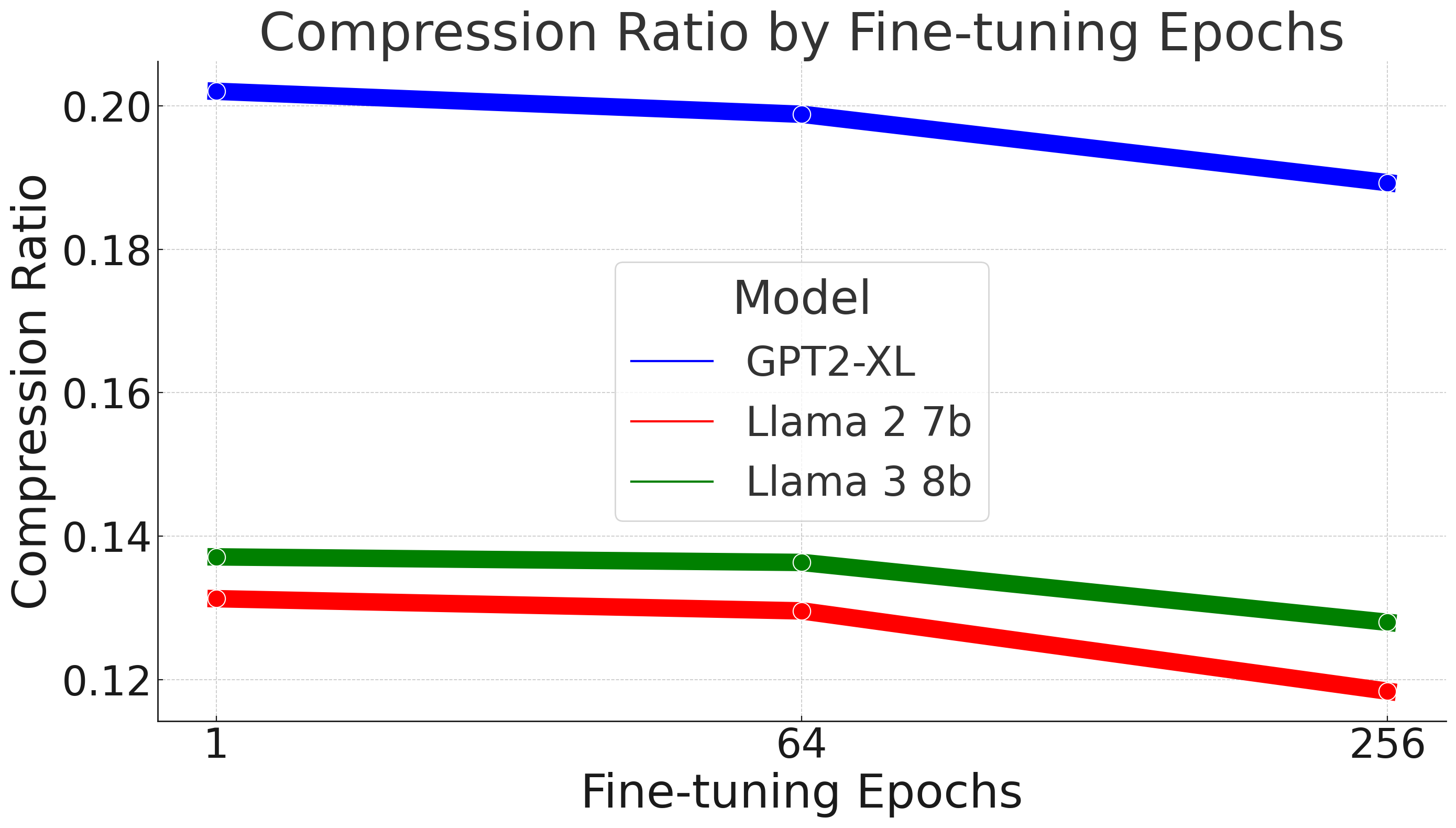} 
    \centering
    \caption{\texttt{FineZip} ablations for different fine-tune epochs}
    \centering
    \label{fig:finetune-lora}
    \centering
\end{figure}

Table \ref{tab:compression_methods} shows that \texttt{FineZip} is able to achieve comparable or better compression ratios than both NNCP and LLMZip with a much faster compression time. Specifically, we see that \texttt{FineZip} has a much better compression ratio than NNCP with comparable amount of compression time, while the 4-bit quantized \texttt{FineZip} is approximately 4 times faster than NNCP and still exhibits a better compression ratio. \texttt{FineZip} compresses enwik8 within 4 hours, compared to approximately 227 hours taken by LLMZip. This is a 54x improvement on compression time with a minor drop of 1 percentage point in compression ratio. 






\subsection{\texttt{FineZip} Ablations}

\texttt{FineZip} uses an "online memorization step" as shown in Figure \ref{fig:system} before performing compression. This is done using Low-Rank Adaptation (LoRA) \citep{hu2021lora}. We compare the effect of fine-tuning on compression using 3 different language models: GPT2-XL 1.3B \cite{gpt-2}, LLama-2 7B \cite{llama2}, and LLama-3 8B \cite{llama3}. We see that for each model, memorization improves the absolute compression ratio by at least 1 percentage point or a relative improvement of about 8\% over its non-fine-tuned baseline as shown in Figure \ref{fig:finetune-lora}. This is significant especially when dealing with such low compression rates. It should be noted that the time taken for memorization is negligible compared to compression time and can be ignored.

\begin{figure}
    \centering
    \includegraphics[width=0.85\linewidth]{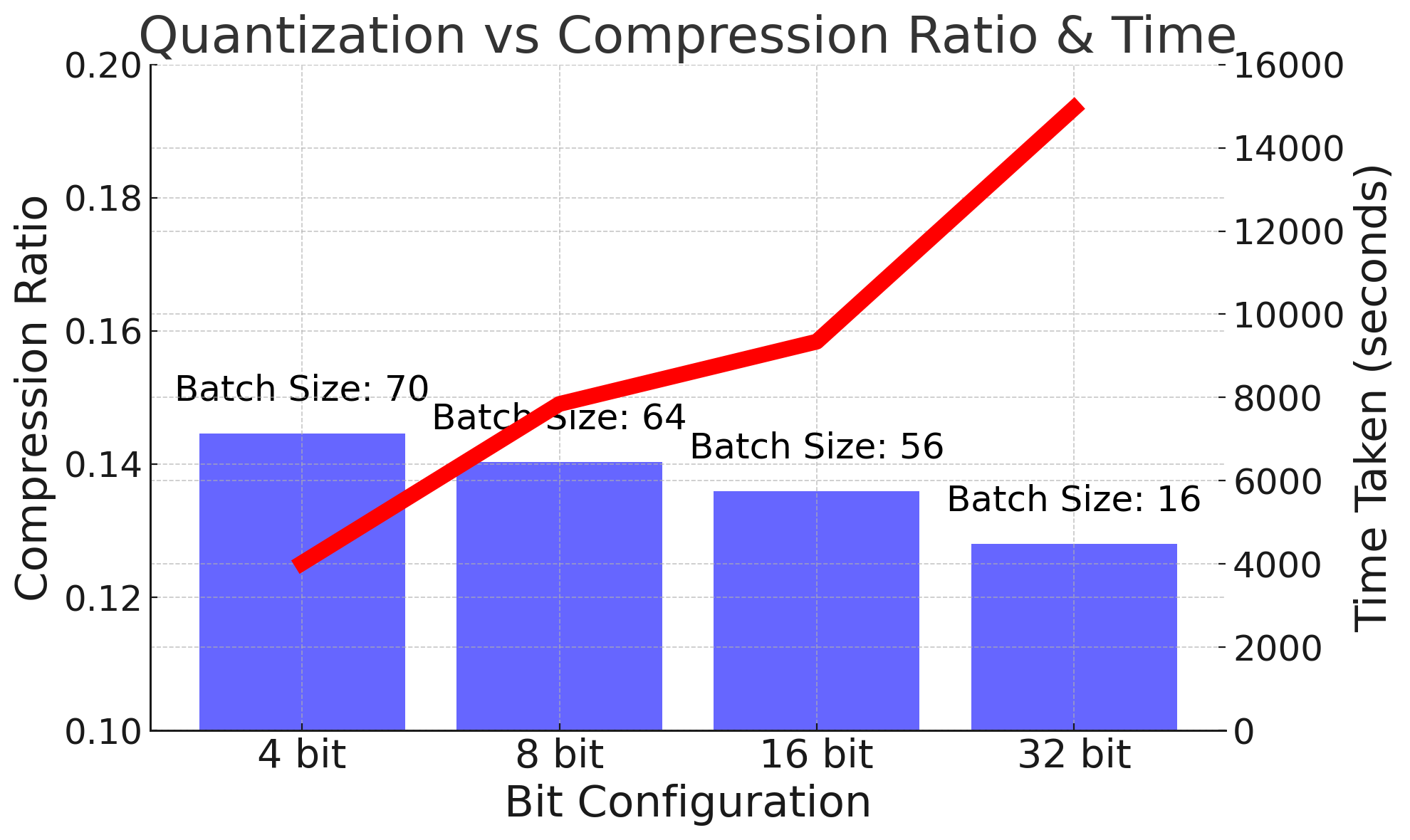} 
    \centering
    \caption{Compressing 10mb dataset with LLama-3 8B loaded with 4, 8, 16, and 32-bit precision. Purple bar shows compression ratio, red line shows time taken to compress. Each batch size was chosen to max out memory on a 48GB GPU.}
    \centering
    \label{fig:quantization}
    \centering
\end{figure}


\paragraph{Quantization:} We saw in Table \ref{tab:compression_methods} that dynamic context helps speed up the compression process by significant amounts, while online memorization is able to mitigate the loss in performance. We further push the limits of compression time using quantization. We perform the memorization step using QLoRA \cite{dettmers2023qlora} and perform compression using the quantized model. We do this using a fixed compute budget of 48GB GPU memory on a single NVIDIA A6000 GPU. Lower precision models will allow us to increase batch size and in turn, decrease time needed to compress a file by a sizable amount. Figure \ref{fig:quantization} shows that fine-tuning/compressing a 4 bit model allows us to fit a batch size of 70 on one A6000 GPU and achieve a compression time of 67 minutes. This 4x speed up makes \texttt{FineZip} not only a competitive compressor out-performning traditional text compression systems by a huge margin, but also the fastest neural network/transformer based compression currently available.

\section{Conclusion}
In this paper we explore the possibility of using LLMs for lossless text compression. We show that while using neural network and LLM based text compression systems lead to significantly better compression rates, they also require impractical amounts of compression time. To alleviate this, we introduce \texttt{FineZip} - an LLM-based lossless text compression system which compresses text 54 times faster than LLMZip with a minor loss in compression performance. \texttt{FineZip} also improves on the compression ratio of traditional text compression systems by approximately 50\%. We also show that while \texttt{FineZip} presents a significant step in making practical text compression systems using LLMs, much still needs to be done. We hope our work can serve as a stepping stone in that direction.

\section{Limitations}
LLM-based text compression systems assume a GPU being available in the host machine for local compression. While this is not true for every personal computer, the landscape is rapidly changing. Many personal laptops are now equipped with GPUs and as compute becomes cheaper and the power of LLMs grow, we envision a future where every personal computer will be equipped with an LLM running locally and performing various tasks.

\bibliography{custom}

\newpage
\appendix

\section{Appendix}

\subsection{Evaluating Traditional Compression Methods}\label{sec:traditional}
\begin{figure}[H]
    \centering
    \includegraphics[width=\linewidth]{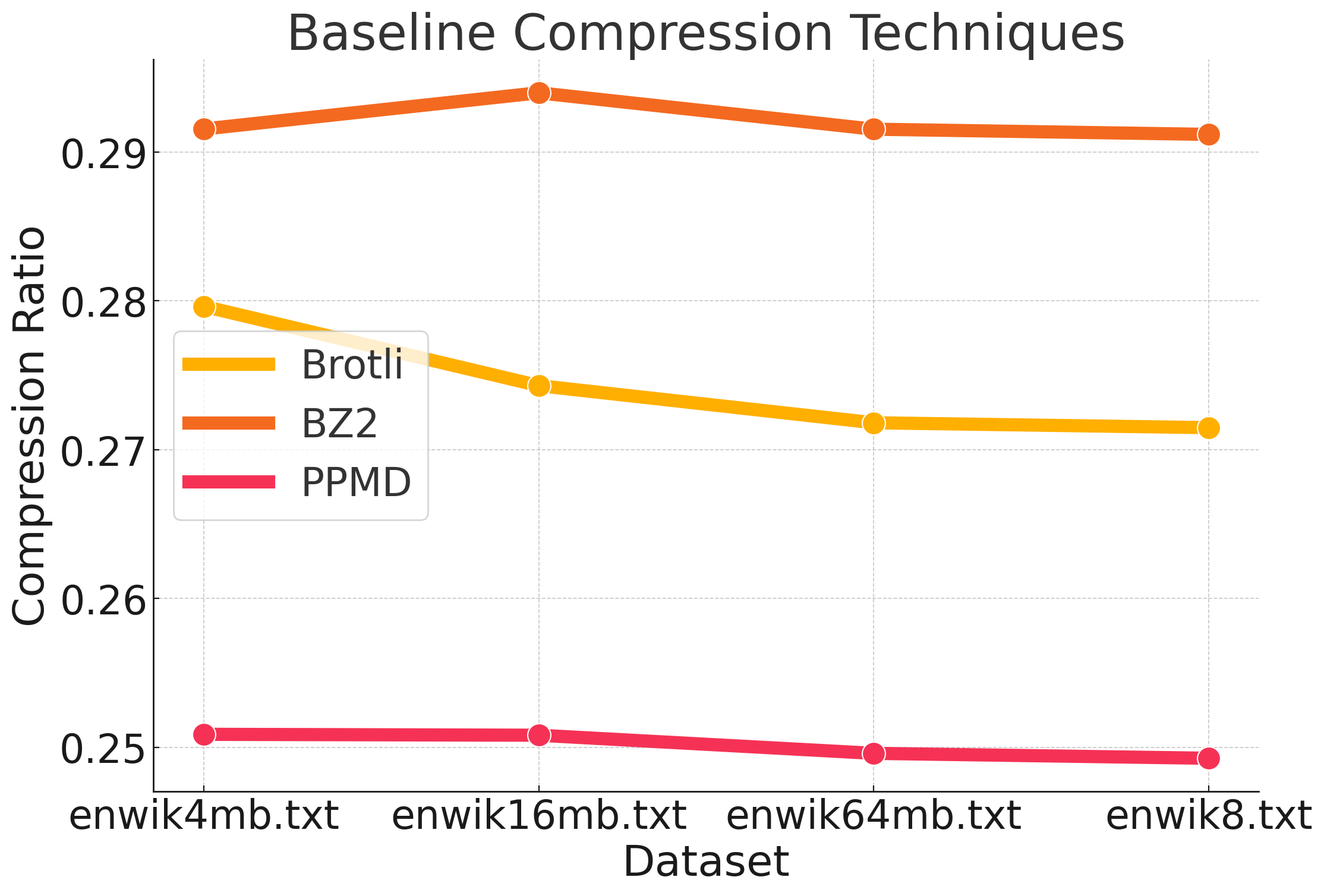} 
    \centering
    \caption{Evaluating Baseline Compression Techniques Brotli, BZ2, and PPM on enwik8}
    \centering
    \label{fig:traditional-text}
    \centering
\end{figure}

We first experimented with three traditional compression methods - Brotli \citep{brotli}, BZ2 \citep{bzip2}, and PPM \citep{1096090} - for text compression as a function of increasing dataset size. We find that PPM performs best for text compression, and that the performance remains relatively constant with respect to dataset size. The results can be seen in Figure \ref{fig:traditional-text}.

We then use these algorithms to compress the ranks generated by LLMs in \texttt{FineZip}. We see that BZ2 has the best performance so we chose it as the traditional compression method for \texttt{FineZip}.

\begin{figure}[H]
    \centering
    \includegraphics[width=\linewidth]{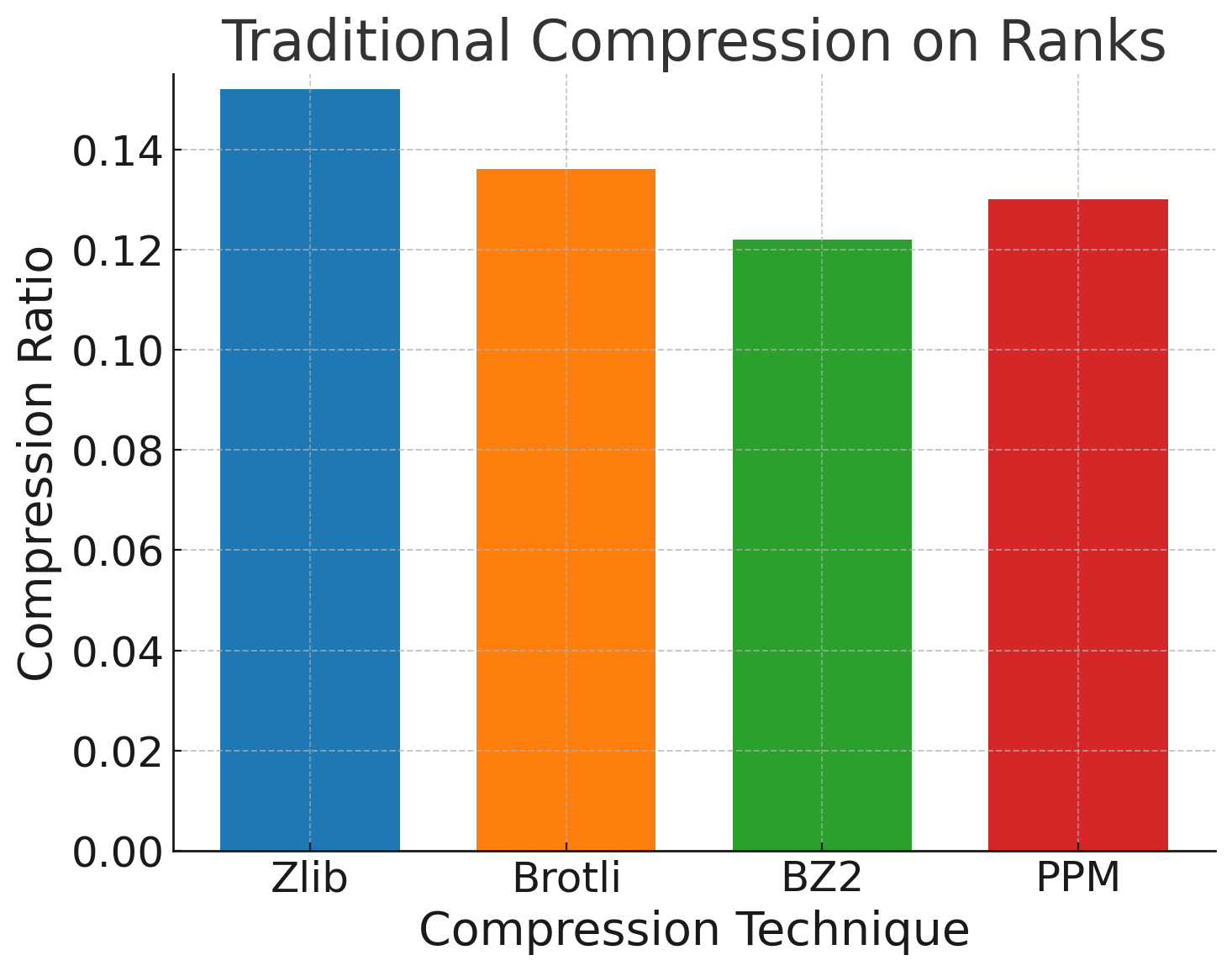} 
    \centering
    \caption{Testing Traditional Compression Techniques Brotli, BZ2, and PPM on the ranks produced by compressing enwik8 with LLama2-7B finetuned for 64 epochs with r=16}
    \centering
    \label{fig:traditional-ranks}
    \centering
\end{figure}

\subsection{Double Compression Benchmark}

Prior to testing \texttt{FineZip}, we compressed the enwik8 \cite{enwik8} dataset using traditional compression techniques (Brotli, BZ2, PPM) to create a benchmark for ourselves. Figure 3 shows that Brotli, BZ2, and PPM perform consistently across varying input file sizes and that PPM performs the best on textual data, reaching a compression ratio of approximately 0.25. Figure 4 measures the compression ratio when two compression techniques are stacked and serves as a more accurate benchmark for \texttt{FineZip} as it also employs two step compression. Through these set of baseline experiments, we can see that a compression ratio of 0.25 is the value to beat. 


\begin{figure}[h]
    \centering
    \includegraphics[width=\linewidth]{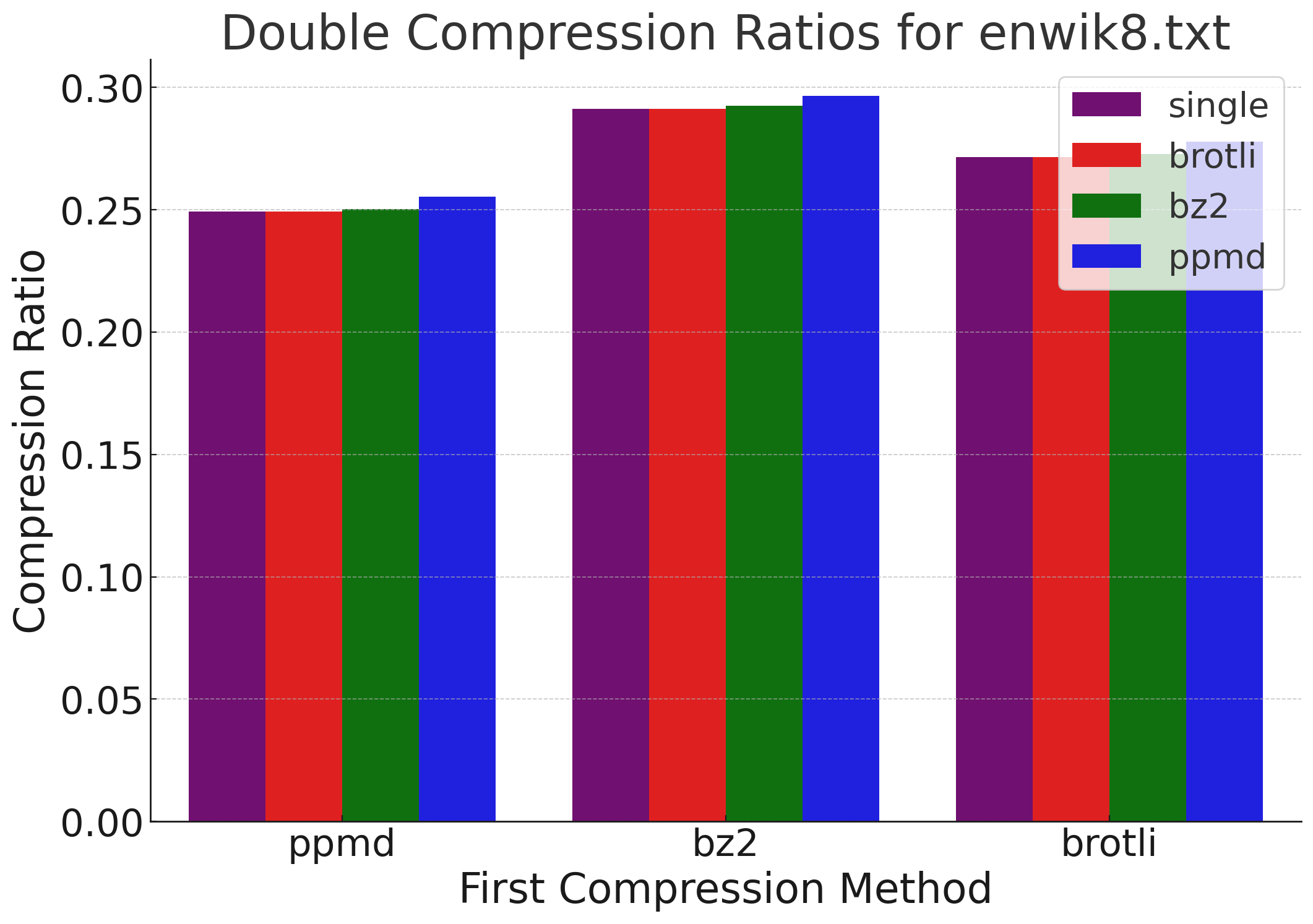} 
    \centering
    \caption{Evaluating Stacked Compression with Brotli, BZ2, and PPM on enwik8}
    \centering
    \label{fig:double-compression}
    \centering
\end{figure}

\subsection{Context Size} \label{sec:context_size}

To determine the best context window size to use, we ran experiments with the LLama2-7B base model (LLMZip) and discovered that a larger context size results in a better compression ratio. The compression ratio began to plateau as the context window reached 512 so we decided to use that for all of our experimentation. 

\begin{figure}[h]
    \centering
    \includegraphics[width=\linewidth]{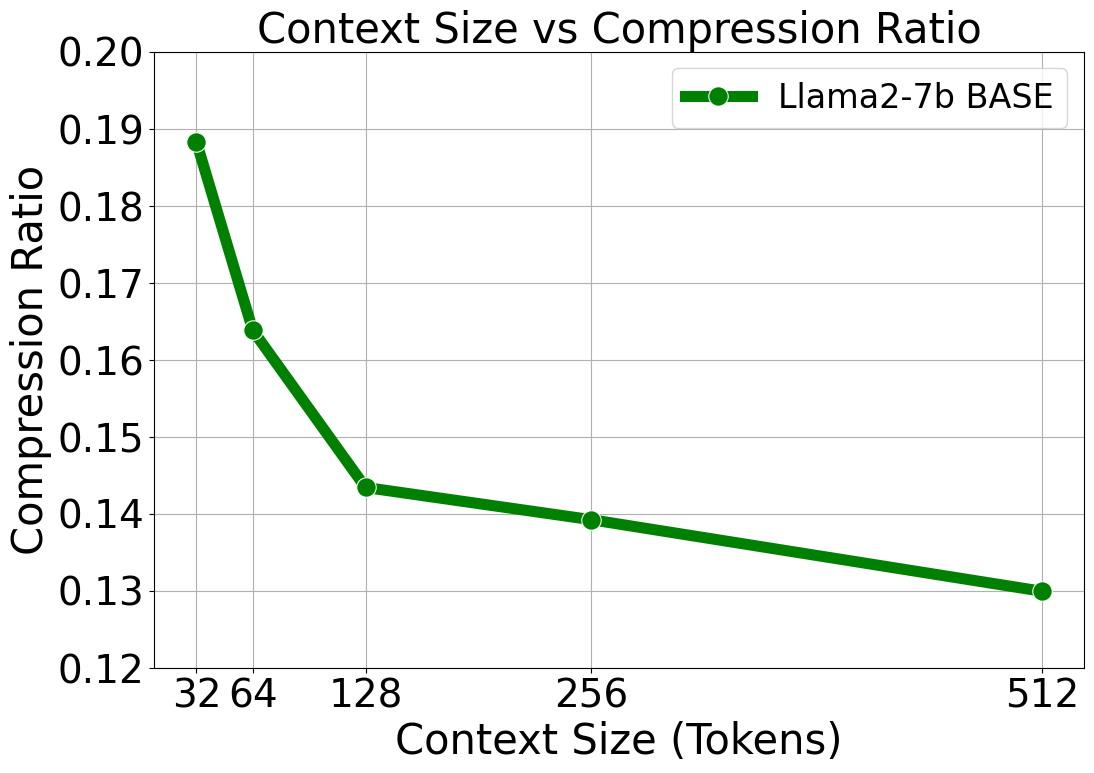} 
    \centering
    \caption{Evaluating Best Context Window for Compression}
    \centering
    \label{fig:double-compression}
    \centering
\end{figure}

\subsection{\texttt{FineZip} and Dataset Size}

\begin{figure}
    \centering
    \includegraphics[width=\linewidth]{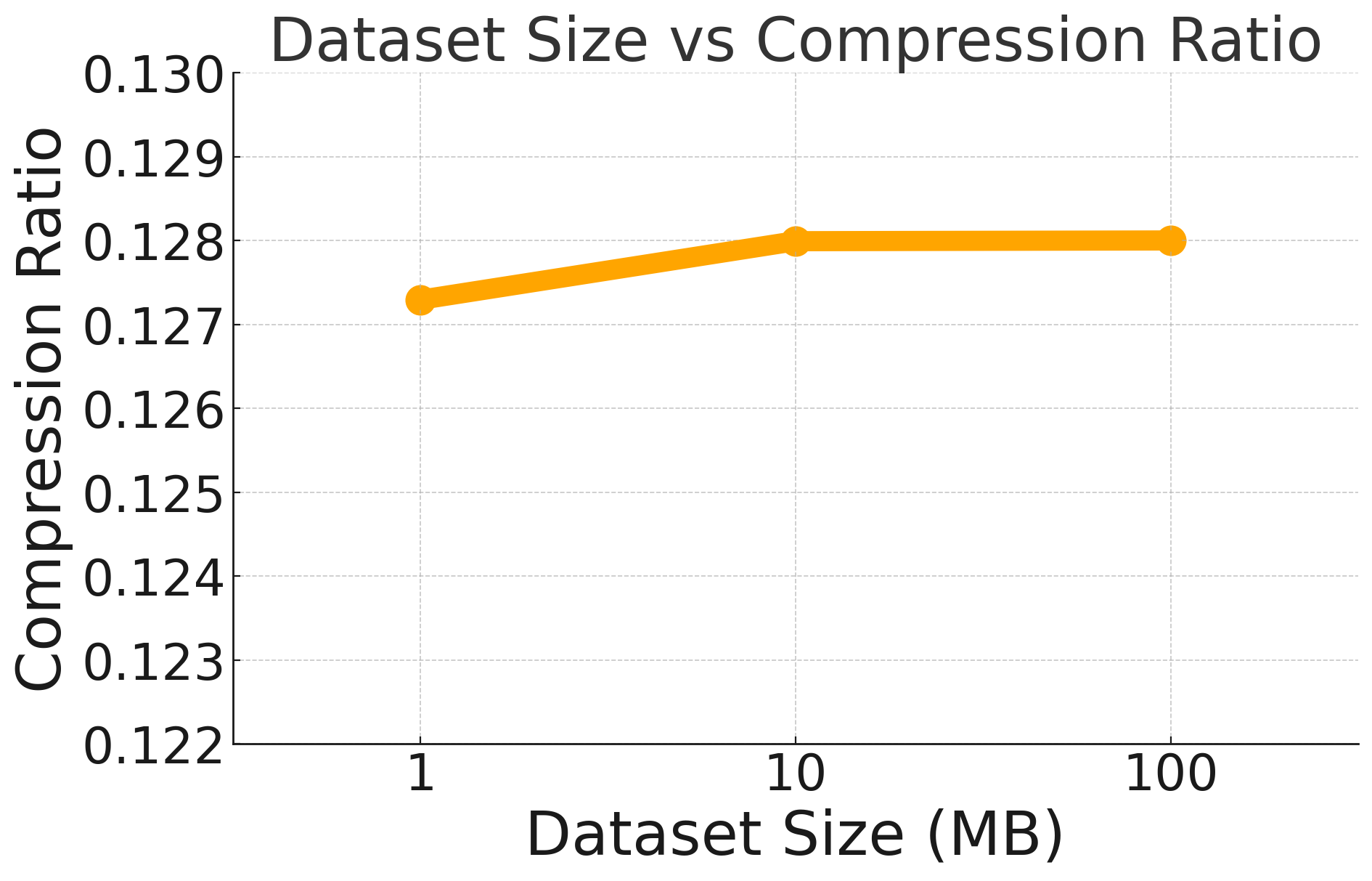} 
    \centering
    \caption{Compressing input files of size 1, 10, and 100 megabytes with LLama-3 8B finetuned for 256 epochs.}
    \centering
    \label{fig:dataset-size}
    \centering
\end{figure}

The previous experiments were only using a dataset size of 10mb and for this to be a viable compression technique, it has to scale well for much smaller and larger file sizes. Figure \ref{fig:dataset-size} shows that LLama-3 8B \cite{llama3} fine-tuned for 256 epochs maintains a consistent compression ratio on dataset sizes of 1, 10, and 100mb. This verifies that \texttt{FineZip} remains viable regardless of dataset size. 

\end{document}